\documentclass[runningheads]{llncs}
\usepackage{graphicx}
\usepackage{pgfplots}
\usetikzlibrary{positioning}

\usepackage[disable]{todonotes} 

\usepackage{pgfplotstable}
\usepackage{booktabs}
\usepackage{times} 

\usepackage[T1]{fontenc}
\usepackage{tabularx}

\usetikzlibrary{arrows}

\usepackage{amsmath}

\usepackage{rotating}

\usepackage{todonotes}
\newcommand{\tod}[1]{\todo[inline]{#1}}
\newcommand{\com}[1]{\todo[inline,color=green]{#1}}
\usepackage[numbers]{natbib}

\usepackage[]{algorithm2e}

\SetCommentSty{mycommfont}

\newcolumntype{C}[1]{>{\Centering}m{#1}}

\tikzset{
  treenode/.style = {align=center, inner sep=0pt, text centered,
    font=\sffamily},
  arn_n/.style = {treenode, circle, white, font=\sffamily\bfseries, draw=black,
    fill=black, text width=1.5em},
  arn_r/.style = {treenode, circle, red, draw=red, 
    text width=1.5em, very thick},
  arn_x/.style = {treenode, rectangle, draw=black,
    minimum width=1.5em, minimum height=1.5em}
}
\begin{document}
\title{Analysis and Optimization of Deep Counterfactual Value Networks}
\titlerunning{Counterfactual Value Networks}  
%
\author{Patryk Hopner \and Eneldo Loza Menc\'{i}a}
\authorrunning{Patryk Hopner \and Eneldo Loza Menc\'{i}a} 
\institute{Knowledge Engineering Group, Technische Universit\"at Darmstadt}

\maketitle              

\begin{abstract}
Recently a strong poker-playing algorithm called DeepStack was published, which is able to find an approximate Nash equilibrium during gameplay by 
using heuristic values of future states predicted by deep neural networks.
This paper analyzes new ways of encoding the inputs and outputs of DeepStack's deep counterfactual value networks based on traditional abstraction techniques, as well as an unabstracted encoding, which was able to increase the network's accuracy.
\end{abstract}
%
%
%
\section{Introduction}
Poker has been an interesting subject for many researchers in the field of machine learning and artificial intelligence over the past decades.
Unlike games like chess or checkers it involves imperfect information, making it unsolvable using traditional game solving techniques.

For many years the state of the art approach for creating strong agents for the most popular poker variant of No-Limit Hold'em involved computing an approximate Nash equilibrium in a smaller, abstract game,
using algorithms like counterfactual regret minimization and then mapping the results back to situations in the real game.
Those abstracted games are however several orders of magnitude smaller than the actual game tree of No-Limit Hold'em, which means, that the poker agent has to treat many strategically different situations as if they were the same, potentially resulting in poor performance.

Recently a work was published, combining ideas from traditional poker solving algorithms with ideas from perfect information games, creating the strong poker agent called DeepStack.
The algorithm does not need to pre-compute a solution for the whole game tree, instead it computes a solution during game play. In order to make solving the game during game play computationally feasible, DeepStack does not traverse the whole game tree, instead it uses an estimator for values of future states. For that purpose a deep neural network was created, using several million solutions of poker sub-games as training data, which were solved using traditional poker solving algorithms.

It has been proven, that, given a counterfactual value network with perfect accuracy, the solution produced by DeepStack converges to a Nash equilibrium of the game. This means on the other hand, that wrong predictions of the network can result in a bad solution. In this paper we will analyze several new ways of encoding the input features of DeepStack's counterfactual value network based on traditional abstraction techniques, as well as an unabstracted encoding, which was able to increase the network's accuracy.

\section{Preliminaries and DeepStack}
The term poker describes a family of card games in which players receive one or more private cards, which
are unknown to the opponents. Players are then betting on whose cards have the highest rank, according
to the rules of the game. 
In this work, we will focus on the most popular poker variant to date, No-Limit (NL) Hold'em in the 2 player variant, called Heads-Up (HU).\footnote{More details on the variants and rules of the game are given by \citet{zinkevich}.} 

Following \citet{zinkevich} poker can be modeled as an \textbf{extensive game}, in which player actions and chance events form a game tree, similar to chess or checkers. The nodes of the tree correspond to \textbf{histories} of actions $h \in H$. Each non-terminal history has an acting player $p(h) \in P$ associated with it, who must select an action $a \in A(h)$. A terminal history $z \in Z \subset H$ has associated utilities $u_i(z)$ for player $i$. 
An information set $I \in I_i$ of player $i$ is a set of histories or states, which can not be distinguished from that player's perspective. 
A player's strategy $\sigma_i \in \Sigma_i$ is a function which assigns a probability to each legal action a player can take in a given information set. 
As all states in one information set are indistinguishable from each other, a player must use the same strategy in all the states in that information set, intuitively this means he must choose a strategy without knowing the opponent's hand distribution. 

A \textbf{Nash equilibrium} \citep{nash} is a strategy profile (the set of all $\sigma_i$) in which no player can increase his expectation by unilaterally deviating from the strategy profile. 
An approximation which loses by at most $\epsilon$ to a best response is referred to as an \textbf{$\epsilon$-Nash equilibrium}. 
The Counterfactual Regret Minimisation (CFR) algorithm \citep{zinkevich} and its variants \citep{public,gibson,tammelin} provably find such an approximation, usually in linear time and space complexity in the number of information sets.
They are considered state of the art algorithms for finding approximate Nash equilibria in imperfect information games and were the basis for the creation of many strong poker bots 
\citep{zinkevich,johanson,schnizlein,tartanian} 
such as  Libratus \citep{libratus} which recently won a competition against human professional players.
The basic idea of CFR is to determine the counterfactual \emph{regret} $R_i(I,a)$ of using the current strategy profile $\sigma$ at an information set $I$ compared to a strategy profile $\sigma_{I\rightarrow a}$ which always plays the action $a$ in this information set.
The regret is computed as the difference $v_i(\sigma_{I\rightarrow a},I)-v_i(\sigma,I)$ between the \emph{counterfactual values} (CV)  of both profiles. 
Roughly speaking, the CV $v_i(\sigma,I)$ corresponds to the average utility of player $i$ when both players play according to $\sigma$ at information set $I$. 
Regrets, CVs, and $\sigma$, which depends on the ratios between the regrets, are iteratively updated and determined in self play, which is guaranteed to converge to a Nash equilibrium \citep{zinkevich}.

However,
whereas the fixed limit variant could be solved using 900 CPU years and 11 Terabytes of Memory \citep{solved}, 
the standard variant in no-limit poker with its $6.3 \cdot 10^{164}$ non-terminal states and $1.39 \cdot 10^{48}$ information sets \citep{size} is still many orders of magnitude too big to be possibly ever solved in an offline manner. 
Hence, a common approach is to compute solutions for abstracted versions of the game. 
The most important abstraction technique is \textbf{card abstraction}, which  is the process of grouping a number of cards together and mapping them to \textbf{buckets}. 
During training cards in the same bucket are considered indistinguishable and only a strategy for each bucket is stored, not a strategy for every individual hand. 
In addition to the usefulness for creating smaller games, card abstractions can also be used to create a feature set for Deep Counterfactual Value Networks (cf. next section), which is the focus of this work. 



\subsection{DeepStack and Deep Counterfactual Networks}
\emph{DeepStack} is a strong poker AI \citep{deepstack}, which combines traditional imperfect game solving algorithms, such as CFR and endgame solving, with ideas from perfect information games, while remaining theoretically sound.

An agent using endgame solving usually plays according to a pre-computed strategy during the first part of the game, called the \emph{trunk}, but computes a solution for the rest of the game, called the \emph{endgame}, during game play \citep{sganzfri15,decomp}.
However, even this technique comes with disadvantages regarding necessary action translations and the off-tree problem, which is why the authors of DeepStack propose to re-solve the sub-tree starting from the current state, after every taken action, instead of using a pre-computed trunk strategy. 
While this continual re-solving offers several benefits over traditional approaches, it is not in itself enough to solve a game as big as NL Hold'em HU, as it would be infeasible to compute a solution in the early stages of the game. For that reason DeepStack introduces \emph{depth limited lookahead}. On the early rounds of the game DeepStack does not traverse the full game tree, but instead uses deep neural networks as an estimator of expected counterfactual values of each hand on future rounds for its re-solving step, resulting in the technique called \emph{depth limited continual resolving}. 

\subsubsection{Deep Counterfactual Value Networks}
DeepStack used a deep neural network to predict the player's counterfactual values on future betting rounds, which would otherwise obtained by applying CFR.
Consequently, the deep counterfactual value network (DCVNN)  is trained with examples consisting of representations of poker situations as input and the counterfactual values of CFR as output.

More specifically,\footnote{More details are given in \citep{supplement}.}
the network was fed with 10 million random poker situations 
and the corresponding counterfactual values 
obtained by applying CFR on the resulting sub-games. 
For every situation a public board, private card distributions for both players and a pot size were randomly sampled.
From this the CFR is able to compute two counterfactual value vectors $\mathbf{v}_i=(v_i(j,\sigma))_j$ with $j=1\ldots1326$ for each possible private hand combination and $i=1,2$ for each player. 
Note that $I=j$ represents the first level of the game tree starting from the given public board.

The input to the network is given by a 
representation of the players' private card distributions and the public cards. 
Hence, before the training of the neural network starts, DeepStack creates a potential aware card abstraction with 1000 buckets, as described in Section~\ref{sec:potentialaware}. 
For each training example the probabilities of holding certain private hands are then mapped to probabilities of holding a certain bucket, by accumulating the probabilities of every private hand in said bucket. 
After the training of the model is completed, the CV for each bucket in a distribution can be mapped back to CV of actual hands by creating a reverse mapping of the used card abstraction.


The network used by DeepStack is a standard feed forward neural network, with 1000 inputs for each player as a representation of his distribution and one additional input for the pot size, for a total of 2001 inputs. The network has 7 fully connected hidden layers with 500 nodes each and uses parametric rectified linear units for the output. 
An additional outer network ensures  the zero-sum property.

DeepStack was able to solve many issues associated with earlier game solving algorithms, such as avoiding the need for explicit card abstraction. 
Moreover, it has been proven, that as the predictions of the deep counterfactual value networks come closer to the true value of a state, depth limited re-solving approaches a true Nash equilibrium of the game. 
However, deep counterfactual value networks introduce their own potential problems though, as wrong predictions of values of future states could potentially result in a highly exploitable strategy. One of the potential reasons for incorrect predictions might be the encoding of the player distributions as well as the counterfactual value outputs. The distributions and the outputs are encoded using a potential aware card abstraction, potentially leading to similar problems as traditional card abstraction techniques, which is something we will call \emph{implicit card abstraction}.

\section{Distribution Encoding}
\label{sec:encodings}

\subsection{Quality of Encodings}
\begin{figure}[t]
    \centering
    \includegraphics[width=0.70\linewidth]{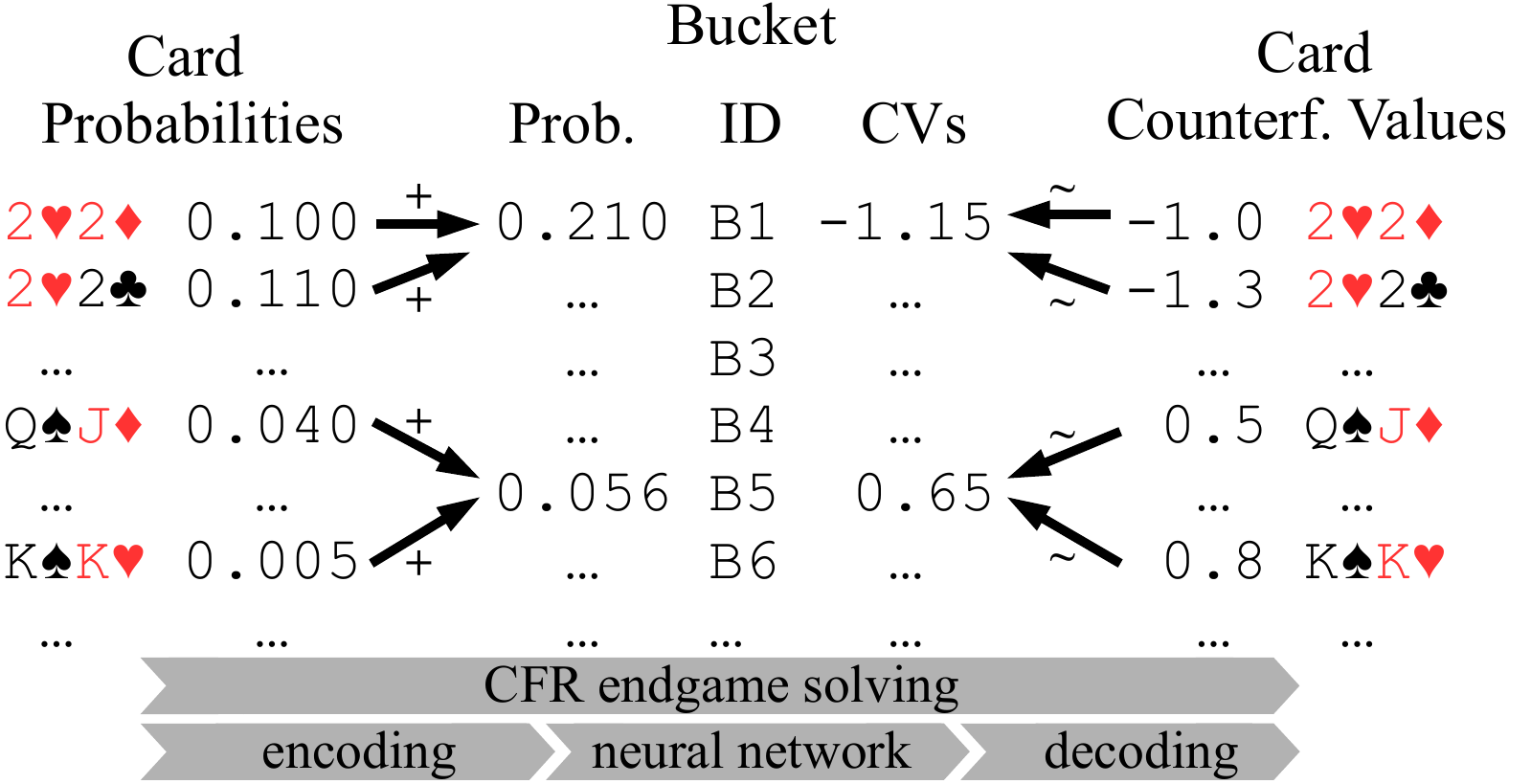}
    \caption{Diagram depicting 1) the encoding from private cards to buckets (depicted by black arrows) 2) the mapping from  private card distributions to bucket distributions (the summing of probabilities symbolized by $+$) 3) the mapping from private card counterfactual values to buckets CVs (averaging symbolized by 
    {\raise.17ex\hbox{$\scriptstyle\mathtt{\sim}$}})
    4) the pipeline of CFR mapping private card distributions to respective CVs 6) the replicating DeepStack pipeline consisting of a) encoding b) estimating of the buckets' CVs by a neural network c) decoding back the estimated buckets' CVs to the cards' CVs.}
    \label{fig:mappings}
\end{figure}
%
While DeepStack never uses card abstraction explicitly during its re-solving step, the encoding of inputs and outputs of counterfactual value networks is based on a card abstraction, which introduces potential problems.

Because the input player distributions get mapped to a number of buckets prior to training, the training algorithm is not aware of the exact hand distributions, but only of the distribution of bucket probabilities. Because this is a many to one mapping, the algorithm might not be able to distinguish different situations, thus not being able to perfectly fit the training set. 

The second problem stems from the encoding of the output values. Counterfactual values of several hands are aggregated to a counterfactual value of a bucket, potentially losing precision. Both problems are visualized in Figure~\ref{fig:mappings} which also depicts the basic architecture of DeepStack's counterfactual value estimation.

While the problem is similar for inputs and outputs, we will focus on the loss of accuracy of counterfactual value outputs when using an abstraction based encoding. We will call the difference between the original counterfactual values of hands, as computed by the CFR solver, and the counterfactual values after an abstraction based encoding was used, the \emph{encoding error}. The difference between the original counterfactual values and the bucket counterfactual values will be measured using the mean squared error as well as the Huber loss (with $\delta=1$) averaged over all private hands and test examples, as proposed by \citep{deepstack}. 
\tod{if space: define Huber loss}
For instance, in Figure~\ref{fig:mappings} we would apply the loss functions on the differences $|-1.0-(-1.15)|$, $|-1.3-(-1.15)|$, $\ldots$ . 
 
We will present three abstraction based encodings, including the potential aware encoding, which was used by DeepStack, as well as an unabstracted encoding. We will then compare the encoding error of each encoding, as well as the accuracy of the resulting networks.

When measuring the accuracy of the model, we have two possible perspectives. The first is to look at the prediction error with both inputs and outputs encoded with a card abstraction.
The second way is to map the predictions of buckets back to predicted counterfactual values of private hands. The predicted counterfactual values of private hands can then be compared to the unabstracted counterfactual values of the test examples. When measuring the error using encoded inputs and outputs, we will refer to the test set as \textbf{abstract test set}. 
In Figure~\ref{fig:mappings} this would correspond to the error between the bucket CVs column (after mapping from the actual private privat card CVs) and the predicted bucket CVs.
When we are measuring the prediction error for unabstracted private hands, we will call the dataset the \textbf{unabstracted test set}, which in Figure~\ref{fig:mappings} corresponds to comparing to the card CVs column after decoding the predicted bucket CVs. 
We will use the same logic for the training set.

\subsection{$E[HS^2]$ Abstraction}
The \emph{hand strength} ($HS$) value of a hand is the probability of winning against a uniform opponent hand distribution on the last betting round the. 
The \textbf{expected hand strength} ($E[HS]$)  on earlier rounds is calculated by averaging the HS values over all possible card roll outs \citep{zinkevich}.

A similar metric is the \textbf{expected hand strength squared} ($E[HS^{2}]$) \citep{johanson} which is obtained by averaging the square of the HS value over all possible card roll outs. While the $E[HS]$ metric only describes the chances of a hand winning at a showdown against a uniform distribution, the $E[HS^{2}]$ metric tries to take the potential of a hand on different rivers into account. The idea is, that a hand which will be very strong on some rivers and weak on others is generally better, than a hand which will be only mediocre on most rivers.
The $E[HS]$ and $E[HS^{2}]$ card abstractions use the $E[HS]$ and $E[HS^{2}]$ values in order to group hands, which are considered to be similar according to the metric, into the same bucket. There are several ways to map hands with given $E[HS]$ / $E[HS^{2}]$ values to a bucket, including percentile bucketing, which creates equally sized buckets, clustering of hands with an algorithm such as k-Means \citep{kmeans} or by simply grouping hands together, that differ only by a certain threshold in their $E[HS]$ / $E[HS^{2}]$ values.

\subsection{Nested Public Card Abstraction}
A \textbf{nested card abstraction} is an abstraction in which hands are first grouped into buckets according to some metric and those buckets are later subdivided. A popular kind of nested bucketing involves \textbf{public card bucketing}. With the approaches described so far, the structure of the public cards was ignored, the metrics focused only on the strength of the private hands. Public card bucketing first divides boards into several groups, allowing the CFR algorithm to store different strategies for each type of board. The buckets are later usually subdivided according to some metric which takes private card information into account, such as $E[HS]$, $E[HS^{2}]$ or a potential aware metric \citep{abs}, which will be described in the next section.

The nested public card abstraction used in this paper first clustered boards into public buckets according to 2 features, the \textbf{draw value} and the \textbf{highcard value}. The draw value of a turn board was defined as the number of straight and flush combinations, which will be present on the following round. The highcard value is the sum of the ranks of all turn cards, with the lowest card, a deuce, having a rank of zero and an ace having a rank of 12. The boards were clustered using the k-Means algorithm and the public buckets were later subdivided by the $E[HS^{2}]$ value of private hands.

\subsection{Potential Aware Card Abstraction}
\label{sec:potentialaware}
While metrics such as the $E[HS]$ value are a good estimate of a hand's general strength, they do not take future potential on different boards into account. The $E[HS^{2}]$ metric tries to measure a hand's future potential more accurately by squaring the $HS$ values of a hand on each river, and in doing so rewarding hands which can develop into a very strong hand and punishing hands which will be mediocre in most situations, but it was only a minor evolution of the $E[HS]$ metric.
The \textbf{potential aware card abstraction} \citep{abs} tries to estimate a hand's potential after future cards more accurately, it does that by first creating a probability distribution of future $HS$ values for each hand and then clustering hands, using the \textbf{k-Means} \citep{kmeans} algorithm and the \textbf{earth mover's distance} \citep{abs}.

In order to obtain the histograms each probability distribution is discretized, with values between 0 and 1, corresponding to the probability of winning against a uniform distribution. All possible future cards are then rolled out, starting from the betting round on which we need to create a bucketing. Once the histograms are created, hands get clustered into a predefined number of buckets, using the earth mover's distance. Since the earth mover's distance measures not only the difference in probability, but also the work that is required in order to transform one histogram into the other, it is better suited for the task of clustering $HS$ histograms, as experiments have also shown \citep{abs}.

\subsection{Abstraction-free Direct Encoding}
Instead of using a card abstraction in order to aggregate private hand distributions to bucket distributions and private hand CVs to bucket CVs, this encoding uses the private hand data directly. 

For the input card distributons all 1326 possible private card combinations will be considered, and the distributions will be encoded as a vector of 1326 probabilities of holding that combination. The board cards will be represented using a one hot encoding: a vector of 52 inputs is used, representing the 52 cards in the deck, if the card is present on the public board, the input is set to one, otherwise it is zero. 

%


\section{Evaluation}
\com{Show here complete comparisons. eg a big table, or two tables, one for abstracted, one ofr unabstracted etc. perhaps also use bar charts: each group of bars is a method, and each method contains eg abstracted and unabstracted. }
\subsection{Experimental Setup}
\com{explain the setup, i.e., how many instances/situations for the encoding error. how many instances for training the network, evaluation on how many. which network used. etc. }
In order to compare the encodings, first a version of each card abstraction described in the previous section was created. Like in the original DeepStack implementation, the potential aware card abstraction used 1000 buckets. 

The $E[HS^{2}]$ abstraction used 1326 equal distance buckets, the possible $E[HS^{2}]$ values, ranging from
zero to one, were divided into 1326 regions and all hands whose $E[HS^{2}]$ values fell into the same region,
were assigned to the same bucket.

The public nested card abstraction was created by first clustering the public boards into 10 public clusters, according to their draw and highcard value and subdividing each public cluster into 100 $E[HS^{2}]$ buckets, resulting in a total of 1000 buckets.

For the analysis of the encoding error, the CVs of each training example were then encoded using each of the 3 card abstractions, meaning that they were aggregated to a CV of their bucket. Those bucket CVs were then compared with the original CVs of the hands in said bucket and the average error over all available training examples was computed.

Unlike the original DeepStack implementation, which used a dataset of 10 million endgame solutions, only 300,000 endgame solutions could be created during the work on this paper. All 300,000 training examples were used for testing the encoding error of each abstarction.

In the second test deep counterfactual value networks were trained using each of the 3 abstraction based encodings, as well as the unabstracted encoding. The training set consisted of 80$\%$ of the total 300,000 endgame solutions, while the test set consisted of 20$\%$. The networks were trained for 350 epochs using the Adam Gradient descent and the Huber Loss.

\subsection{Encoding and Prediction Errors}
\begin{table}[t]
\centering
\caption{Encoding error of different encoding schemes on the turn.}
\label{tab:encodingerror}
\begin{tabular}{|c|c|c|c|}
\hline
\textbf{\begin{tabular}[c]{@{}l@{}}Encoding Approach $\backslash$ \\ Encoding Error \end{tabular}} &
\textbf{$E[HS^2]$} &
\textbf{\begin{tabular}[c]{@{}l@{}}Public Nested\\ \end{tabular}} &
\textbf{Potential Aware} 
 \\ \hline
\textbf{Huber loss} & 
\textbf{0.0240} & 
0.0406 & 
0.0258  
\\ \hline
\textbf{MSE}       & 
\textbf{0.0509} & 
0.0886 & 
0.0544  
\\ \hline
\end{tabular}
\end{table}

Table~\ref{tab:encodingerror} shows the encoding error of mapping the hand distribution to the respective encoding, and back. 
Table~\ref{tab:predictionerror} reports the errors of the trained neural networks. 
Remember that the abstraction-free encodings do not produce any encoding error, therefore, their performance is also the same on the abstracted and unabstracted sets.
Note also that the errors on the abstracted sets are not directly comparable to each other due to the different encoding.

Regarding the $E[HS^2]$ abstraction, we can observe that it  
 introduces a smaller encoding error than the potential aware card abstraction, although not by a big margin. 
However, it is outperformed in terms of the accuracy of the neural networks. The potential aware abstraction performed better in its own abstraction, as well as after mapping the counterfactual values of buckets back to counterfactual values of cards.

A somehow contrary behaviour can be observed for the public nested encoding. 
Whereas it has major difficulties in encoding, the resulting encodings carry enough information for the network to predict relatively well on the bucketed CVs. However, mapping the CVs back to the actual hands strongly suffers from the initial encoding problems.


However, the most noteworthy 
(and surprising) 
result is the performance of the abstraction-free encoding. 
Whereas the potential aware encoding was able to produce a lower Huber Loss in its own abstraction, the abstraction-free encoding outperformed the abstraction on the unabstracted training set and the unabstracted test set. The direct encoding was therefore better than the potential aware encoding at predicting counterfactual values of actual hands instead of buckets, which is the most important measure in actual game play.
These results suggest that the neural network was able to generalize among the public boards even though no explicit or implicit support was given in this respect. 
Note that this was possible even though 
we only used a small number of training instances compared to DeepStack.

\begin{table}[t]
\centering
\caption{Prediction error of neural network using different input encodings on the abstracted and unabstracted train and test sets, on the turn.}
\label{tab:predictionerror}
\begin{tabular}{|c|c|c|c|c|}
\hline
\textbf{\begin{tabular}[c]{@{}l@{}}Encoding Approach $\backslash$ \\ Huber Loss \end{tabular}} &
\textbf{$E[HS^2]$} &
\textbf{\begin{tabular}[c]{@{}c@{}}Public Nested \\ \end{tabular}} &
\textbf{\begin{tabular}[c]{@{}c@{}}Potential \\ Aware \\ \end{tabular}} &
\textbf{\begin{tabular}[c]{@{}c@{}}Abstraction--\\ Free \\ \end{tabular}} 
 \\ \hline
\textbf{Abstracted Train} & 
0.0254 & 
0.0080 & 
\textbf{0.0052} & 
0.0102    
 \\ \hline
\textbf{Unabstracted Train} & 
0.0387 & 
0.0436 & 
0.0267 & 
\textbf{0.0102}    
 \\ \hline
\textbf{Abstracted Test} & 
0.0330 & 
0.0161 & 
\textbf{0.0102} & 
0.0143   
 \\ \hline
\textbf{Unabstracted Test} & 
0.0434 & 
0.0478 & 
0.0297 & 
\textbf{0.0143}   
\\ \hline
\end{tabular}
\end{table}


\section{Conclusions}
In this paper we have analyzed several ways of encoding inputs and outputs of deep counterfactual value networks. We have introduced the concept of the encoding error, which is a result of using an encoding based on lossy card abstractions. An encoding based on card abstraction can lower the accuracy of training data by averaging counterfactual values of multiple private hands, introducing an error before the training of the neural network even started. We have observed, that the encoding error can have a substantial impact on the accuracy of the trained network, as observed in the case of the public nested card abstraction, which performed well on its abstract test set, but lost a lot of accuracy when the counterfactual values of buckets were mapped back to hands.

The potential aware card abstraction produced the best results of all the abstraction based encodings, which corresponds to the results achieved by the abstraction in older algorithms, where it is the most successful abstraction at this point.

The unabstracted encoding produced the lowest prediction error. While a good result on the training set might be expected, it was unclear if the neural network would generalize well to unseen test examples. This result again shows the importance of minimizing the encoding error, when designing a deep counterfactual value network.

\renewcommand{\bibname}{References}
\footnotesize
%
%
{}

\end{document}